# Deep Neural Networks and Neuro-Fuzzy Networks for Intellectual Analysis of Economic Systems

**Alexey Averkin[a] and Sergey Yarushev[b]**

[a] Laboratory Dorodnicyn Computing Centre, Faculty, University, Vavilov st. 40, 119333, Moscow, Russia, Plekhanov Russian University of Economics, Stremyanny lane, 36, Moscow, 117997, Russia e-mail: averkin2003@inbox.ru

[b]Department of Informatics, Faculty, Plekhanov Russian University of Economics, Stremyanny lane, 36, Moscow, 117997, Russia, e-mail: Yarushev.SA@rea.ru

**Abstract**

In tis paper we consider approaches for time series forecasting based on deep neural networks and neuro-fuzzy nets. Also, we make short review of researches in forecasting based on various models of ANFIS models. Deep Learning has proven to be an effective method for making highly accurate predictions from complex data sources. Also, we propose our models of DL and Neuro-Fuzzy Networks for this task. Finally, we show possibility of using these models for data science tasks. This paper presents also an overview of approaches for incorporating rule-based methodology into deep learning neural networks.

## 1 Introduction

Many machine learning tools can be effectively wielded by analysts to reduce the computational burden on an analyst [2]. One such machine learning tool, deep learning (DL), looks to encode complex mathematical representations using stacked auto encoders. One popular algorithm within DL is convolutional neural networks (CNNs), which have proven their utility in object classification [6] and detection [8] within imagery. CNNs can be used to process large amounts of data and sort out what and where an object is located without requiring analysts to manually sort through the imagery. While this is a useful tool, there exists a breakdown in communication between the operator and the CNN. The CNN is able to accurately generate a classification label but does not necessarily report on features that were present allowing a classification to be inferred. For example, a CNN may be able to correctly identify an object as being a 'cat' but not have any representation of 'whiskers' or 'fur.' Similarly, the analyst would not be able to communicate the importance of a specific feature or trait to the CNN which limits the amount and nature of feedback from an analyst to a CNN.

This problem fundamentally limits the utility of such tools. Without understanding how a CNN arrives at a solution, it is impossible to understand how adaptable the system is. This poses a complex challenge for many autonomous systems, as many of these machine learning tools are developed with controlled imagery and trained on labeled data. However, when an autonomous system is deployed the imagery may fundamentally change and there are no guarantees that the machine learning tools will operate effectively given these changes.

The paper deals with the task of building a hybrid time series forecasting system based on deep neural networks and cognitive modeling. This approach allows us to take into account both the quantitative and qualitative characteristics of the time series. For completeness, the features of fuzzy cognitive maps and their application in problems of time series forecasting are given. Also, the developed genetic algorithm for learning fuzzy cognitive maps is presented, which allows to avoid the time-consuming task of manually setting up a cognitive map. To solve the problem of working with semistructured data, which often take place in the tasks of forecasting time series, it is proposed to use deep neural network architectures, since such networks are able to operate with this type of data and show the most reliable results.

## 2 Neuro-Fuzzy networks in forecasting tasks

ANFIS is the abbreviation Adaptive Neuro-Fuzzy Inference System - an adaptive network of fuzzy output. Proposed in the early nineties [4], ANFIS is one of the first variants of hybrid neural-fuzzy networks - a neural network of direct signal propagation of a special type. The architecture of the neural-fuzzy network is isomorphic to the fuzzy knowledge base. Neuro-fuzzy networks use differentiated implementations of triangular norms (multiplication and probabilistic OR), as well as smooth functions. This allows the use of cross-fuzzy neural networks, rapid algorithms for learning neural networks, based on the method of back propagation of errors. The architecture and rules for each layer of the ANFIS network are described below. ANFIS implements the Sugeno fuzzy inference system in the form of a five-layer neural network of direct signal propagation [7].

The network inputs in a separate layer are not allocated Figure 1. shows an example of an ANFIS network with two input variables (x1 and x2) and four fuzzy rules. In the example, the linguistic evaluation of the input variable x1, three terms are used, and for the variable x2 are used two terms.

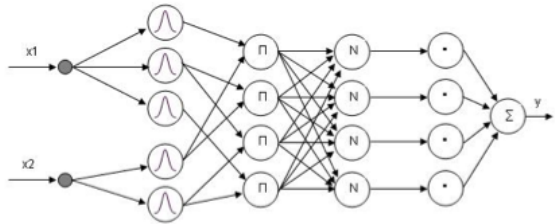

Figure 1: ANFIS structure.

Typical procedures for learning neural networks can only be used to configure the ANFIS network. Usually, a combination of gradient descent is used in the form of an algorithm for back-propagation of errors and least squares parameters. The error back-propagation algorithm configures the parameters of the rules antecedents, i.e. membership function. The method of least squares estimates the coefficients of the concluded rules, since they are linearly related to the output of the network. Each procedure is performed in two stages. At the first stage, a learning choice is introduced, and there is a relationship between the desired and actual network behavior; the iterative least squares method is the optimal parameters of the nodes of the fourth layer. The method of back propagation of errors changes the parameters of the nodes of the first layer. It is aligned with the first stage of the coefficients. The iterative tuning procedure continues until a predetermined value is set. For example, the Levenberg-Marquardt method.

To date, there are a large number of different approaches to the neuro-fuzzy forecasting of time series, in particular, to study networks using genetic algorithms, methods of swarm optimization, as well as various hybrid methods. Chinese scientists [9] presented their own model of forecasting financial flows in the banking sector, called APAPSO (adaptive activity in the field of population, PSO). Scientists have proposed a hybrid learning algorithm based on the APAPSO algorithm using the least squares method. In comparative experiments using the comparison algorithm with the standard back-propagation error method in combination with the minimal squares method (LMS), as well as with the traditional swarm optimization method-LMS.

In order to avoid a problem, it is necessary that the algorithm be calculated on the basis of the PSO algorithm. Using APAPSO algorithm in accordance with the diversity of the population and the speed of movement of particles. In this algorithm, the PSO algorithm is improved, which allows optimizing the diversity of particles and possibilities for expanding the algorithm from local optimal solutions.

The results demonstrated an increase in the rate of optimization, as well as a forecast of growth accuracy.

Scientists from the USA and Iran [1] presented a joint work on research in the field of forecasting the dollar / yen currency pair. In their work, researchers compare the performance of ANFIS models with previously developed models of user neural networks. Mediterranean RMS (Table 1) and results are better for ANFIS than for Sugeno Yasukawa, ANN or several regression approaches. As can be seen from the table, the mean square error is the least in models based on ANFIS.

| FORECASTING MODEL | RULES | ITERATIONS | RMSE |
|---|---|---|---|
| Multiple regression | - | 10 | 7,9221 |
| Neural network | - | 10 | 2,8082 |
| sugeno-yasukawa | 6 | 10 | 4,8290 |
| ANFIS | 7 | 10 | 2,6301 |

Table 1: Comparison of forecasting methods

The researchers concluded that although the difference between the traditional ANN and the ANFIS system is small, the ANFIS system is a much more human-readable system and more convenient to use.

Indian scientists [3] proposed a hybrid forecasting model based on the integration of ANFIS and an immune algorithm for predicting the Indian stock market. To create an effective prediction model, scientists decided to use an artificial immune algorithm to tune the parameters of the membership function of a fuzzy inference system. As an input for testing the system, data of daily trading closures on the National Stock Exchange of India (NSE) were used, as well as well-known technical indicators. At the exit, a forecast of the future value of the NSE index was obtained.

Algorithm of clonal selection. An artificial immune algorithm is a new method of intelligent computing based on the human biological immune system. In this study, scientists used an immune algorithm called clonal selection. This algorithm works like a genetic algorithm. It simulates natural biological cells. Antibodies are fixed on B maple, which recognizes antigens coming from the external environment. Also, this algorithm clones more antibodies with the best-fit antibodies to destroy the antigen. As a result, the immune system produces more antibodies than antigens. The basic elements in this algorithm are antibodies and antigens. In ANFIS, antigens are input to the stock market, and antibodies are unclear rules. The algorithm is used to train fuzzy rules with the best fit to the stock market data.

The results of the experiments were compared with other models based on soft calculations and actual data from the auction. As a result, the results of the experiment showed that the proposed prediction model gave much more accurate prediction results in comparison with traditional models.

Despite the large number of diverse approaches to network training, modification of the architecture and hybridization of several methods, the ANFIS model lies at its core. Each study demonstrated the superiority of the neuro-fuzzy approach over traditional neural networks and other statistical methods.

## 3 Deep neural networks in forecasting tasks

Deep learning neural networks is based on teaching perceptions and not on specialized algorithms designed for specific tasks. Many deep learning methods were known as early as the 1980s, but the results were unimpressive [10], while advances were made in the theory of artificial neural networks (pre-training of neural networks using a special case of an undirected graphical model, the so-called limited Boltzmann machine) x (above all, Nvidia GPUs, and now Google's tensor processors) did not allow creating complex technological architectures of neural networks with sufficient performance yu and allow to solve a wide range of tasks, do not be an effective solution before, for example, in computer vision, machine translation, speech recognition, with quality solutions, in many cases, are now comparable, and in some cases superior to "the protein" experts. Unlike machine learning, depth learning requires a much larger amount of training sample than in the case of machine learning. Also, unlike machine learning, a deep neural network can have thousands of layers. All this helps deep neural networks to achieve a sufficiently high accuracy in the tasks of analysis, classification and image recognition. But, the main drawback of in-depth training is the enormous resource-intensiveness; in order to train a deep neural network, it is sometimes necessary to make a training sample of a million images, or even more, and the learning process can take several days. For such tasks, even developed separate GPU processors to speed up the learning process.

To date, there are a number of libraries for deep neural networks. The most popular of them are Tensor Flow and Keras, which can be used in forecasting problems. In the work of scientists from Australia [5] Keras is used for short-term forecasting of energy consumption in the private sector. This task is very difficult, since the indicators in this case vary greatly. Scientists have concluded that prediction using a deep neural network allows to obtain acceptable results.

### 3.1 Hybrid system for time series analysis

The forecasting system is based on a modular architecture, which gives additional stability to the system, even if one of the modules fails, the other modules continue to do their work.

The system itself has three main modules responsible for the task of forecasting. The deep neural network performs a time series forecast based on numerical indicators and gives us a so-called quantitative forecast, the results of which pass through a verification system (assessment of forecast adequacy), probability of its fulfillment, that is, with a consonance factor that tells us whether the forecast will be fulfilled or not. Further, all data obtained from these modules come to the third module, working on the basis of the neuro-fuzzy network, which aggregates the information obtained from the previous modules and gives the final forecast. In Figure 2 is a diagram of a forecasting system.if the forvecast corresponds to the required accuracy, it is transmitted to the next module. In parallel with the deep network, the module works with a fuzzy cognitive map, which receives input data on the eventual impact on the time series, a cognitive map is constructed, which takes into account all factors influencing a specific predicted indicator. At the output, a cognitive map gives us a forecast with the probability of its fulfillment, that is, with a consonance factor that tells us whether the forecast will be fulfilled or not. Further, all data obtained from these modules come to the third module, working on the basis of the neuro-fuzzy network, which aggregates the information obtained from the previous modules and gives the final forecast. In Figure 2 is a diagram of a forecasting system.

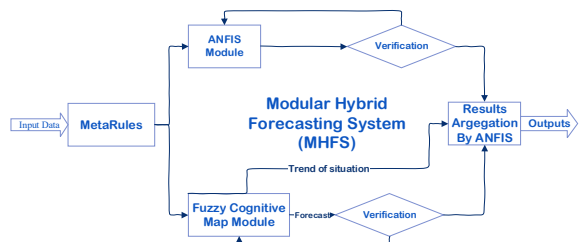

Figure 2. Modular forecasting system

## Acknowledgement

This research was performed in the framework of the state task in the field of scientific activity of the Ministry of Science and Higher Education of the Russian Federation, project "Development of the methodology and a software platform for the construction of digital twins, intellectual analysis and forecast of complex economic systems", grant no. FSSW-2020-0008.